\documentclass[conference, a4paper]{IEEEtran} 

\usepackage[ruled,norelsize]{algorithm2e}
\makeatletter
\newcommand{\removelatexerror}{\let\@latex@error\@gobble}
\makeatother
\ifCLASSINFOpdf
	\usepackage[pdftex]{graphicx}
	\renewcommand{\figurename}{Figure}
\else
\fi
\begin{document}

\IEEEoverridecommandlockouts
\IEEEpubid{\begin{minipage}{\textwidth}\ \\[12pt] %\centering
\\\\
  Personal use of this preprint copy is permitted. Republication, redistribu-\\
  bution and other uses require the permission of IEEE.\\
  DOI: 10.1109/IWSSIP.2016.7502762, \copyright 2016 IEEE.\\
\end{minipage}} 
% \IEEEoverridecommandlockouts
% \IEEEpubid{\makebox[\columnwidth]{
% 978-1-5090-4429-0/17/\$31.00 \copyright 2017 IEEE \hfill} \hspace{\columnsep}\makebox[\columnwidth]{ }}

\title{Segmented and Directional Impact Detection\\ for Parked Vehicles using Mobile Devices}

\author{\IEEEauthorblockN{Andr\'{e} Ebert, Sebastian Feld, and Florian Dorfmeister}
\IEEEauthorblockA{Mobile and Distributed Systems Group\\
Institute for Computer Science\\
Ludwig-Maximilians-University, Munich\\
Oettingenstr. 67, 80538 Munich, Germany\\
Email: \textit{andre.ebert@ifi.lmu.de, sebastian.feld@ifi.lmu.de, florian.dorfmeister@ifi.lmu.de}}}


%Telephone: (800) 555--1212, Fax: (888) 555--1212}


\maketitle

% \fi
\IEEEpeerreviewmaketitle

\begin{abstract}
Mutual usage of vehicles as well as car sharing became more and more attractive during the last years. Especially in urban environments with limited parking possibilities and a higher risk for traffic jams, car rentals and sharing services may save time and money. But when renting a vehicle it could already be damaged (e.g., scratches or bumps inflicted by a previous user) without the damage being perceived by the service provider. In order to address such problems, we present an automated, motion-based system for impact detection, that facilitates a common smartphone as a sensor platform. The system is capable of detecting the impact segment and the point of time of an impact event on a vehicle's surface, as well as its direction of origin. With this additional specific knowledge, it may be possible to reconstruct the circumstances of an impact event, e.g., to prove possible innocence of a service's customer. 
\end{abstract}

\begin{IEEEkeywords}
% Computer Society, IEEE, IEEEtran, journal, \LaTeX, paper, template.
collision detection, templates, inertial sensors
\end{IEEEkeywords}

\section{Introduction}\label{sec:introduction}

With new business models like car sharing there are also new problems arising in the automotive sector: when renting a vehicle it may already be damaged, but because of low-grained maintenance intervals this may not be registered by the service provider. Moreover, it is vague who has to come up for damages or complications resulting from malfunctioning vehicles.
In that context, automated localization and temporal assignment of collisions may help prove the innocence of a stakeholder. Furthermore, automated impact detection may support maintenance personal and enabling more efficient work flows on the service provider's side.
Commonly, heavy collisions usually cause severe and easily visible damages; moreover the vehicle may not be operational anyways. That is why the emphasis of our work is the detection of impact events with low impact forces, which are probably not immediately visible to the human eye. To enable a possible reconstruction of the circumstances of an accident, we also want to identify the direction, an impact originated from. In order to keep the system as simple as possible and to minimize its costs, we employ only one common smartphone as a sensor platform. 
\par
After presenting previous relevant work in Section \ref{sec:relatedwork}, we explain our concept for segmented and directional detection of impacts on parked vehicles in Section \ref{sec:concept}. It identifies an impact's location on predefined segments of the vehicle and its direction of origin, exclusively on basis of our template-based motion data analysis approach. Moreover, we implemented an automated live analysis for continuous event data streams. For the prototypical setup, we used an RC-car as a test vehicle. Subsequently, we evaluate the system's performance for our three different template-based approaches in Section \ref{sec:evaluation} and sum up our findings as well as open issues in Section \ref{sec:conclusion}.

%%%%%%%%%%%%%%%%%%%%%%%%%%%%%%%%%%%%%

\section{Related Work}\label{sec:relatedwork}
% In the given context, there are several approaches based on evaluation of vehicle-driven data gathered with a smartphone, which can be divided into two categories: collision detection and driving behavior recognition.
In context of motion data analysis for vehicles White et al. present WreckWatch \cite{white2011wreckwatch}, a system for automated detection of accidents. 
They evaluate acoustical and acceleration data to create a tuple model consisting of 11 resettable conditions, e.g., loudness, x-,y-, and z-acceleration, etc., which is analyzed by their algorithm.
This approach contains valuable ideas for our intentions, though it is not capable of detecting collisions on a specific vehicle segment and its usage of audio data may be disturbed by ambient sounds. 
A concept for recognizing aggressive and non-aggressive driving maneuvers is provided by Johnson et al. \cite{johnson2011driving}. 
Therefore, they use acceleration data, the euler rotation, and gravity information. 
Moreover, they create classes of maneuvers and split them into longitudinal and lateral movements (e.g., right or left turn, aggressive right or left).
Afterwards, they compare the distances of the low pass filtered output signals with in prior created templates via Dynamic Time Warping (DTW) and reach a rate of 97\% of correctly classified aggressive maneuvers \cite{berndt1994using}. 
Engelbrecht et al. provide a similar approach \cite{engelbrecht2014recognition}, but with less accuracy.
Another template-based concept with the goal of detecting drunk driving under the usage of acceleration and orientation data is proposed by Dai et al. \cite{dai2010mobile}.
Fazeen et al. created a system to increase safe driving by alerting the driver in hazardous situations \cite{fazeen2012safe}, where information about bumpers, potholes, lane changes, etc. is extracted. 
SenSpeed by Han et al. \cite{han2014senspeed} estimate the speed of vehicles by using acceleration data and angular velocity, whereas 85\% of its estimation errors are lower than $5mph$. 
Wang et al. determine the position of a mobile phone within a driving vehicle on basis of motion data analysis and try to predict its usage by the driver \cite{wang2013sensing}.
In the field of human motion analysis, Lih et al. developed a fall detector for elderly people \cite{kau2014smart}. It analyzes a simplified energy model and is based on the sum of acceleration data combined with gyroscope output. It reaches 92\% of correctly classified falls. Kostopoulos et al. are analyzing human fall accidents by utilizing a threshold-based algorithm  achieving a sensitivity of 93.5\% and a specificity of 98.5\% \cite{kostopoulos2015f2d}.

%%%%%%%%%%%%%%%%%%%%%%%%%%%%%%%%%%%%%

\section{Concept for segmented impact detection}\label{sec:concept}
Prerequisite for the following concept is the assumption, that pushes from different directions onto a vehicle create individual rotation and acceleration patterns, which may be used to identify the impact location as well as the direction of origin of a push.

\subsection{Defining a Segmented Impact Model}\label{subsec:collisionmodeldefinition}
\begin{figure}[!t]
\centering
\includegraphics[width=2.5in]{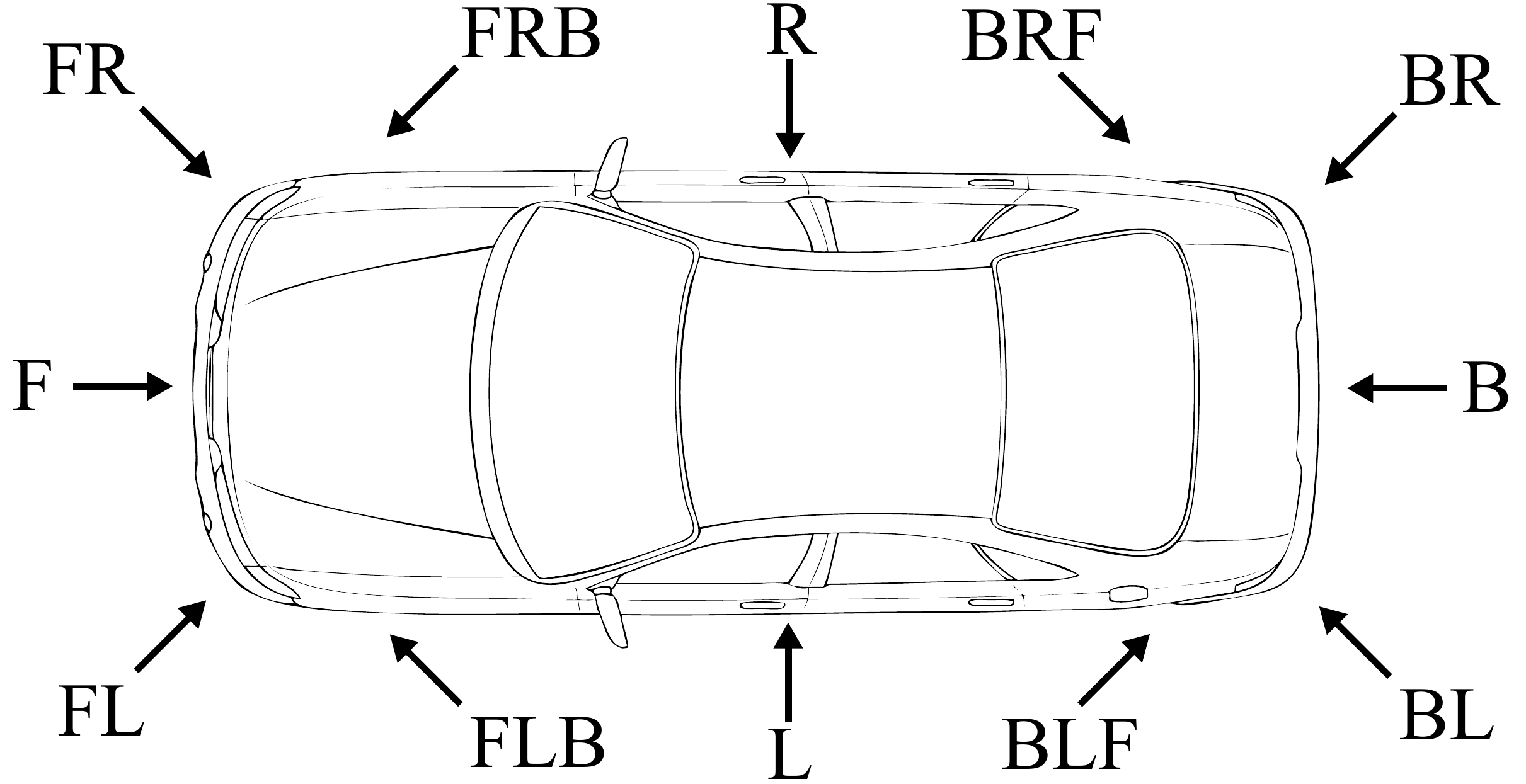}
\caption{12 defined impact segments, diversified into 3 impact categories: Straight Impacts (F,B,L,R), Diagonal Impacts (FL, FR, BL, BR), Reversed Diagonal Impacts (FLB, FRB, BLF, BRF).}
\label{fig:pushingdirections}
\end{figure}

To classify pushing directions and impact segments onto a vehicle's surface, we determined 12 different classes of collisions (see Fig. \ref{fig:pushingdirections}) which can be diversified into 3 different impact categories. Reason for that are similar motion and rotation patterns, occurring for all classes within a category, respectively: \textbf{(1)} \textit{Straight Impacts} are caused by straight pushes with an angle of $90^\circ$ onto the vehicle's surface. Classes within that category are front (F), back (B), left (L), and right (R).
\textbf{(2)} \textit{Diagonal Impacts} are caused by diagonal pushes onto the vehicle's corners with an angle of $45^\circ$. Classes within that category are front-left (FL), front-right (FR), back-left (BL), and back-right (BR). \textbf{(3)} \textit{Reversed Diagonal Impacts} are caused by diagonal pushes onto the vehicle's corners, but conducted from the contrary directions ($135^\circ$) compared to diagonal impacts. The identified classes are front-left pushed from the backside (FLB), front-right pushed from the backside (FRB), back-left pushed from the front (BLF), and back-right pushed from the front (BRF).

\subsection{Acquiring Motion Sensor Data}\label{subsec:acquiringmotionsensordata}
To acquire motion data for further analysis, we implemented an Android App, running on a Samsung Nexus 3 as a sensor platform. For our prototypical test setup, we adjusted the device with a lace onto a radio controlled vehicle, whereby the accelerometer's y-axis $a_{y}$ proceeds along the vehicle's forward driving direction, its x-axis $a_{x}$ represents lateral movement, upward and downward motions are captured by the z-axis $a_{z}$ with the display turned upwards. The RC-car has a weight of $2.155kg$ and its wheels and suspension are supporting a dynamic oscillation during collision events. This allows us to verify our concept without the occurrence of high monetary costs due to damages on real world cars.
Tracked data is the vehicle's acceleration $A =\{a_{x}, a_{y}, a_{z}\}$, its rotation $R = \{r_{x},r_{y},r_{z}\}$ and corresponding time stamps with a sampling rate of $100Hz$. The data is saved locally and streamed into a network via User Datagram Protocol (UDP) additionally.  A Java application listens within the network and performs a live event analysis.

\subsection{Data Preprocessing}\label{subsec:datapreprocessing}
To smooth our data and to reduce signal noise, we designed a Butterworth low pass filter with a cutoff of $f_{c}=20Hz$ to target frequencies higher than those representing the collision events, which where determined to be in a range of $2-10Hz$.

\subsection{Event Extraction and Processing}
\label{subsec:algorithmiceventextraction}
Before template creation, we recorded sample data for each defined impact segment. We conducted 50 pushes per segment and evoked an average acceleration of $4m/s^2$ - $8m/s^2$. The pushes were evoked by pushing the test vehicle manually by hand. Subsequently, we started a serial event detection process in order to detect the single collisions within the data. Therefore, we designed a specific, threshold-based peak detection algorithm. The threshold was determined by analyzing our sensor's noise for 5 minutes. The detected noise maximums were ~$0.9m/s^2$ and $-0.9m/s^2$, so we used a threshold of $1.0m/s^2$ and $-1.0m/s^2$. This  means, a minimum force of $2.155N$ is needed to trigger the event detection, disregarding friction forces. A peak is defined to be the first (and commonly strongest) local positive or negative extrema. It indicates the entry point into a collision event. Algorithm \ref{alg:eventdetection} depicts the event frame creation via peak  detection schematically.
\begin{figure}[!t]
 \removelatexerror
  \begin{algorithm}[H]
  \label{alg:eventdetection}
  
	\SetKwFunction{algo}{algo}\SetKwFunction{detectPeak}{detectPeak}
	\SetKwFunction{algo}{algo}\SetKwFunction{analyzeData}{analyzeData}
	\SetKwFunction{algo}{algo}\SetKwFunction{createEventFrame}{createEventFrame}
  
  	\tiny
  	\DontPrintSemicolon
  	\caption{Serial, continuous peak detection}
   
   	$A{y}=\{a_{y1},..,a_{yn}\}$; // \textit{y-axis acceleration data}\; 
   	$A{x}=\{a_{x1},..,a_{xn}\}$; // \textit{x-axis acceleration data}\; 
   	% $lastYPosVal, lastXPosVal = 0$; // \textit{last positive acceleration values} \;
   	% $lastYNegVal, lastXNegVal = 0$; // \textit{last negative acceleration values} \;
   
   	\;
   
   	%% Analysis function
   	\analyzeData{$A_{y}, A_{x}$}
   	\;
	\For{$c := 1$ to $n$}{
		$isYPeak, lastYPosVal, lastYPosVal =$ \;
	 		\detectPeak($A_{yc}, lastYPosVal, lastYNegVal$);\;
		\If{($isYPeak$)}{
			$isXPeak, lastXPosVal, lastXNegVal =$ \;
	 		\detectPeak($A_{xc}, lastXPosVal, lastXNegVal$);\;
		}
		\If{($isYPeak$ \textbf{and} $isXPeak$)}{
			\createEventFrame(); // \textit{create event frame of  500ms length}\;
			$c = c + createdFrameLength$; // \textit{skip next 500ms}
		}
	}

   %% Peak detection function
   \;
   \detectPeak{$a_{c}, posVal, negVal$}{
   		\;
		\uIf {($a_{c} >= posThresh$)}{
			\eIf {($a_{c} >= posVal$)}{
				$posVal = A_{c}$;
			}{
				$isPeak =$ \textbf{true}; // \textit{entry point of collision event detected}
			}
		}
		\uElseIf{($a_{c} < posThresh$ \textbf{and} $posVal > 0$)}{
				$isPeak =$ \textbf{true}; // \textit{entry point of collision event detected}
		}
		\;

		\uIf {($a_{c} <= negThresh$)}{
			\eIf {($a_{c} <= negVal$)}{
				$negVal = a_{c}$\;
			}{
				$isPeak =$ \textbf{true}; // \textit{entry point of collision event detected}
			}
					
		} 
		\uElseIf{($a_{c} > negThresh$ \textbf{and} $negVal < 0$)}{
				$isPeak =$ \textbf{true}; // \textit{entry point of collision event detected}
				
		}
   	\textbf{return} $isPeak, posVal, negVal$;\;
   		
   }
   
  \end{algorithm}

\end{figure}
Input are two continuous streams of acceleration data along the x- and y-axes. Within the \verb+analyzeData()+ function, the acceleration along the y-axis is checked serially for peaks by \verb+detectPeak()+. If both acceleration values indicating a peak at the same time, a $500ms$ enduring event ($50ms$ padding plus $450ms$ upcoming values) is cut out of the input data and framed for further analysis. The average length of an event was determined empirically.
% With the events framed into these time windows, they are now ready for further processing, e.g., template creation or classification.

\subsection{Template Creation}\label{subsec:templatecreation}
Subsequently, we propose three different approaches for template creation: Approach 1) is a serial DTW distance comparison using a set which contains the 10 most similar templates for the acceleration vectors $A_{y}$, and $A_{x}$ out of all recorded test events per impact segment, 2) is 1 template for $A_{y}$, $A_{x}$, and the rotation $R_{z}$, covering the mean values of this set, 3) is 1 median template for $A_{y}$, $A_{x}$, and $R_{z}$ also created from the single values of the first approach's set. For Pruning, we measured the DTW distances of all templates to each other and split them into two clusters with the k-Means algorithm. Only the cluster with the smaller value distribution is taken for template creation in order to avoid overfitting. Prerequisite for this advance is the assumption that a greater value distribution in one of the clusters is created by outliers and less relevant representatives.

\subsection{Segmented and Directional Impact Detection}
After the creation of templates, all prerequisites for the detection and classification of impact events are met. Our algorithm iterates serially through given test data and detects impact events first. Therefore, the algorithm described in Section \ref{subsec:algorithmiceventextraction} is used again. After their detection, extracted events become handled over to the decision unit and depending on the facilitated template approach (see Section \ref{subsec:templatecreation}) the DTW distance ratios to the given templates are calculated for their x-acceleration as well as their y-acceleration: $D_{xr} = \frac{D_{xmin}}{D_{xmax}}$, $D_{yr} = \frac{D_{ymin}}{D_{ymax}}$. The ratios are used instead of absolute values to make them comparable and to avoid overweighting of single vectors. Finally, the template set with the lowest vector norm $V_{n} = \sqrt[]{D_{xr}^2 + D_{yr}^2}$ indicates the direction, as well as the segment where a push onto a vehicle's surface came from. In the second experiment, where we also used the z-axis' rotation ratio $D_{zr} = \frac{D_{zmin}}{D_{zmax}}$ within the decisioning process, the vector norm is calculated as follows: $V_{n} = \sqrt[]{D_{xr}^2 + D_{yr}^2 + D_{zr}^2}$.

% \subsection{Live Analysis}\label{subsec:liveanalysis}
% The proposed system is designed to manage a live data analysis. Due to serial iteration through input data fixes, it can operate with in prior recoded data sets as well as with live data packages streamed via UDP.  

%%%%%%%%%%%%%%%%%%%%%%%%%%%%%%%%%%%%

\section{Evaluation}\label{sec:evaluation}
In this section, our system design is reviewed regarding its capabilities of successful collision classification as well as its runtime and other parameters. The second set of testdata, which contains 611 collision events (609 detected, see Section \ref{subsec:miscanalysis}) was analyzed with all 3 templating approaches for 12 impact segments. We processed low pass filtered as well as raw data and all in all we completed 3654 classifications. 

\subsection{Acceleration-based with multiple templates per segment}\label{subsec:accelerationdata}
\begin{figure}[!t]
\centering
\includegraphics[width=3.8in]{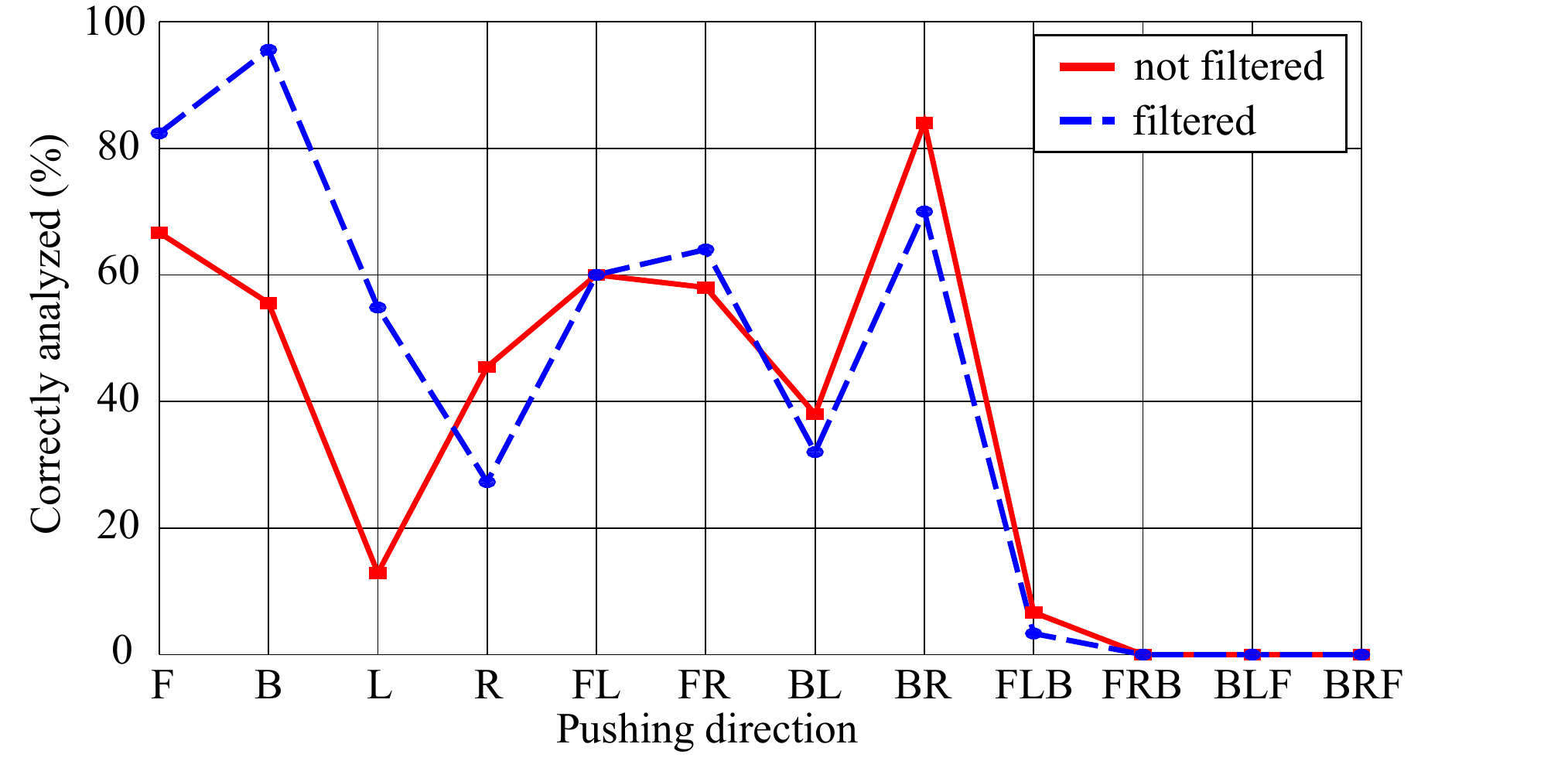}
\caption{Results for acceleration-based approach with 10 templates per segment}
\label{fig:serialperformance}
\end{figure}

\begin{figure}[!t]
\centering
\includegraphics[width=3.55in]{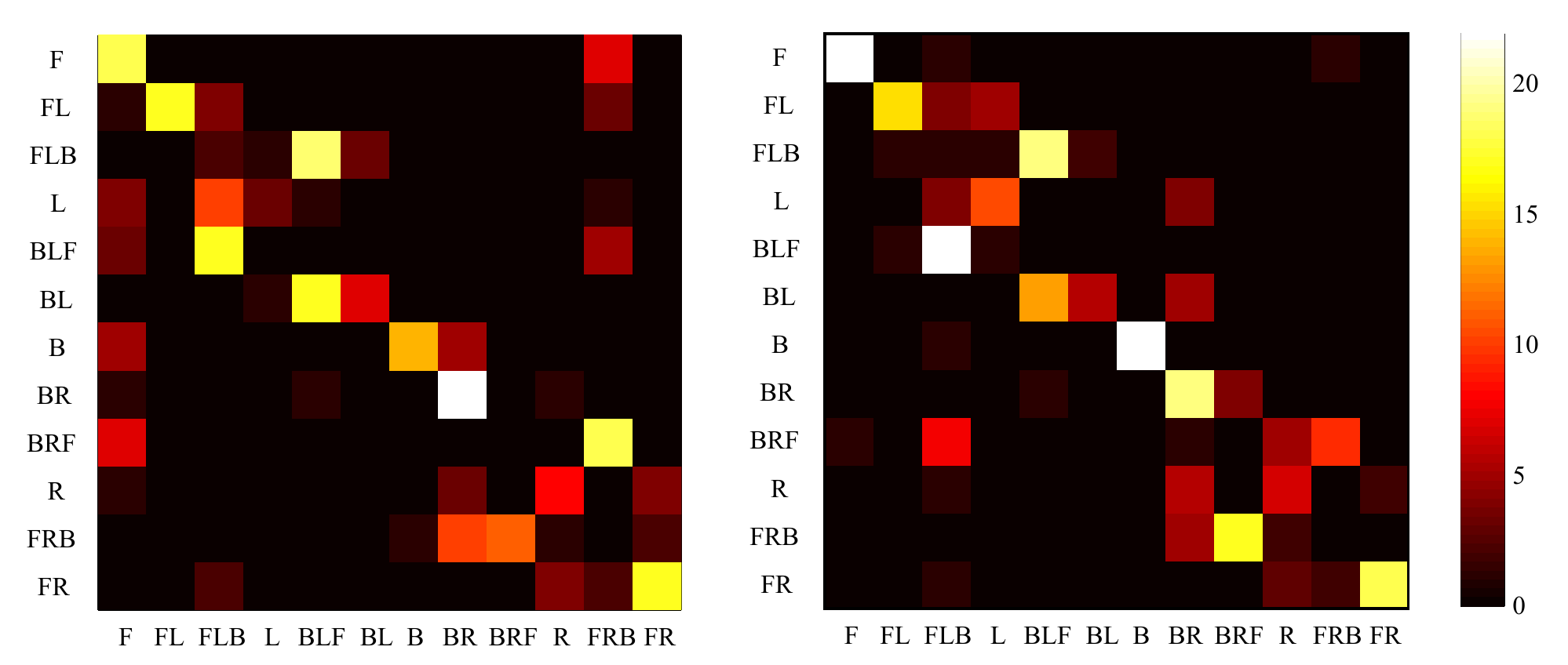}
\caption{Detected impact segments ordered in $360^\circ$ around the test vehicle for the 10 most similar templates approach, not-filtered (left) and filtered (right).}
\label{fig:serialrelvals}
\end{figure}

The system's performance for the usage of the 10 most similar templates per impact segment is visualized in Fig. \ref{fig:serialperformance}. We reached an average rate of correct classifications of only 35.4\% without low pass filtering, and 39.6\% using a low pass.
Fig. \ref{fig:serialrelvals} shows the actual impact segments in relation to the segments our system assigned to them. The diagonal pattern shows that wrongly classified pushes were often assigned to segments, which are neighbors of the groundtruth's segment.

\subsection{Combined analysis with mean and median templates}
In order to improve the correct classification rate of the prior approach, we additionally included the vehicle's rotation information for the z-axis into our analysis. To speed up the systems runtime, we merged all 10 templates (for each motion vector and each impact segment, respectively) to 1 single template by layering them and describing them first by their mean (2) and second by their median (3) values.
\label{subsec:accelerationandrotation}
\begin{figure}[!t]
\centering
\includegraphics[width=3.4in]{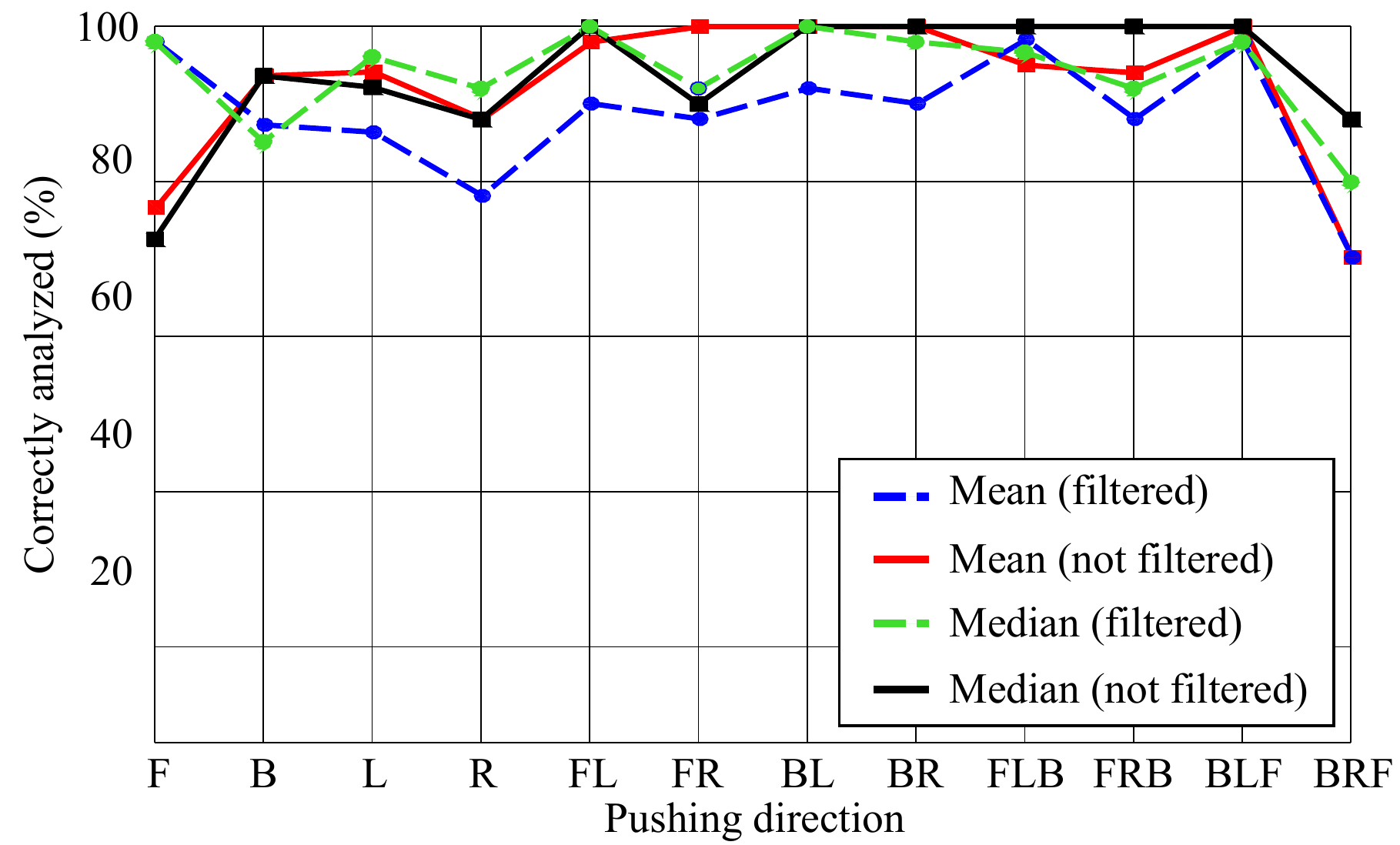}
\caption{Results for acceleration- and rotation-based approach with 1 filtered (f) and 1 not-filtered (nf) mean and median template per segment.}
\label{fig:medianandmeanperformance}
\end{figure}
Fig. \ref{fig:medianandmeanperformance} shows, that these changes improved the system's performance significantly. The average success rate with unfiltered mean templates (2) is 92.4\%, with an additional low pass filter it is 88.8\%.
The median approach accomplishes the best results with an average success rate of 94\% for not filtered, and 94.1\% for filtered data. These results show, that $r_{z}$ contains significant information for impact classification as well as for identifying reversed diagonal pushes. They were not detectable in (1), but now they discovered with an average success rate of nearly a 100\%.
\begin{figure}[!t]
\centering
\includegraphics[width=3.5in]{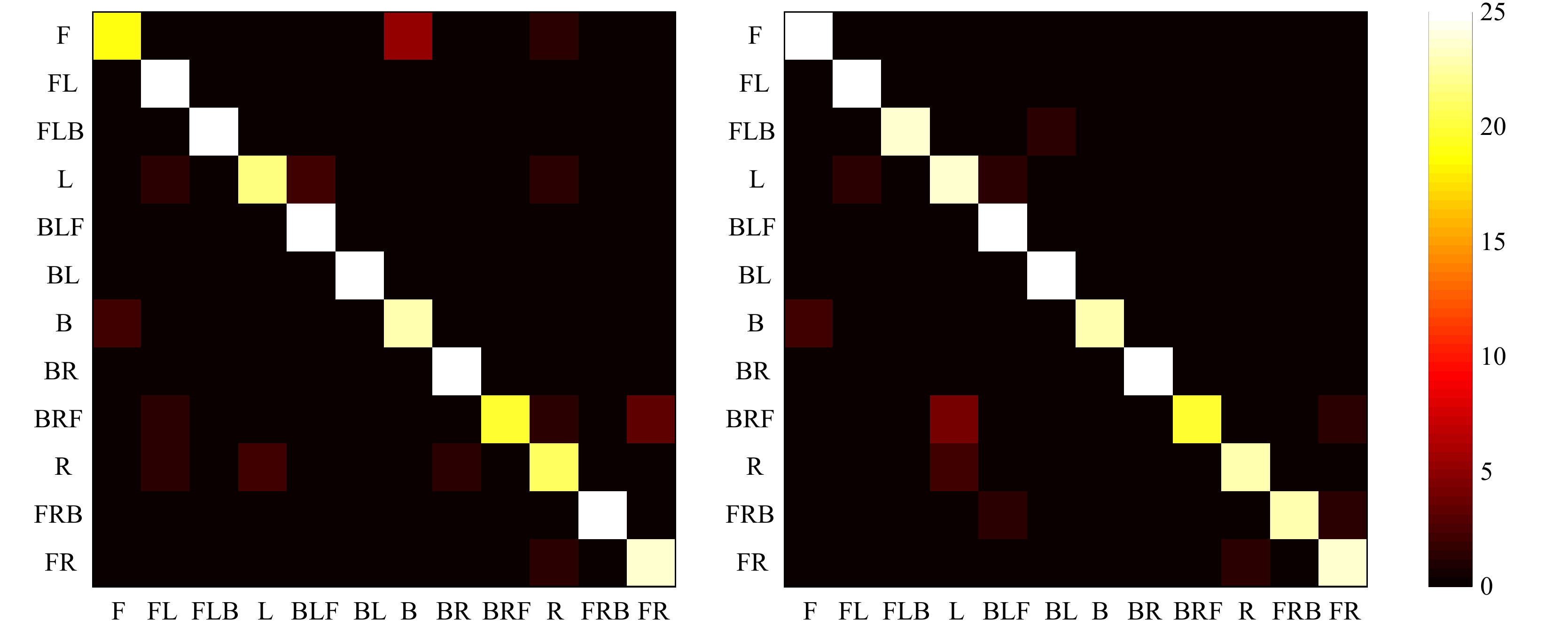}
\caption{Detected impact segments for median templates, not-filtered (left), and filtered (right).}
\label{fig:medianrelvals}
\end{figure}
Fig. \ref{fig:medianrelvals} shows the actual impact segments in relation to the computed segments for median templates. Due to filtering, the classification errors for the R-segment are reduced, although the error for the BRF-segment is enlarged.

% \begin{figure}[!t]
% \centering
% \includegraphics[width=3.55in]{graphics/rel_vals_dist_both_mean.pdf}
% \caption{TODO}
% \label{fig:meanrelvals}
% \end{figure}
% mean vs. median, no serial because of much to long runtime 

\subsection{Runtime}\label{subsec:runtime}
Table \ref{tab:runtime} shows the elapsed time during the analysis in relation to the number of analyzed events. Using multiple templates within the first approach was significantly slower than mean (2) or median (3) approaches. (2) was the fastest with $1.7s$ for analyzing one collision event, though it was slightly less accurate than (3). Both, mean and median, are suitable for a slightly delayed live data analysis. Thereby, an extra of $500ms$ needs to be added to the analysis frame to capture the time until an active event has finished and its extraction may start.

\begin{table}[htb]
\scriptsize
\caption{Runtime for each approach in relation to the quantity of analyzed events.}
\label{tab:runtime}
\vspace{-2mm}

\begin{tabular}{llll}
\textbf{Approach} & \textbf{Events} & \textbf{Runtime} & \textbf{Time / Event}
\\\hline\vspace{-1.5mm}
\\(1) 10 Templates ($A_{y},A_{x}$) & 1218 & $13681s$ & $11.2s$\\
(2) Mean Templates ($A_{y},A_{x},R_{z}$) & 1218 & $2116s$ & $1.7s$ \\
(3) Median Templates ($A_{y},A_{x},R_{z}$) & 1218 & $2688s$ & $2.2s$ \vspace{0.5mm} \\
\hline
\end{tabular}

\end{table}

\subsection{Additional notes}\label{subsec:miscanalysis}
The main reason for incorrect classifications within the category of straight pushes (F,B,L,R) is the force which acts along the axis orthogonal to the vehicle's impact segment. In this case, it is small and hence does not create a distinct acceleration pattern. Still, it is weighted as strong as all other motion vectors with distinct patterns. This problem may be addressed algorithmically by ignoring orthogonal accelerations for straight pushes (e.g., low orthogonal forces) or a more specific usage of filtering methods. Not taken into account before, but also notable is the average success rate of the event detection algorithm. Reason for that is its affection on the system's overall success. With 608 correctly detected collisions and 1 falsely detected event in a dataset of 911 events, the average overall success rate for the event detection algorithm is more than 99\%.

%%%%%%%%%%%%%%%%%%%%%%%%%%%%%%%%%%%%%

\section{Conclusion}\label{sec:conclusion}
We presented a concept for segmented and directional impact detection for parked vehicles with one mobile device. We explained the different steps of data preprocessing, event extraction, template creation, and event analysis. The three proposed template-based approaches have different performances regarding correct classification rate and runtime for the 12 different impact segments and varying pushing directions. We achieved average success rates of more than 94\% and our system is suitable of providing shortly delayed live data analysis. Due to additional analysis of rotation data, we are also able to determine contrary pushing directions and angles from each other, distinctly. Still, our system is not able to provide a progressive analysis without impact segments. Moreover, there are open issues concerning the weighting of orthogonal signals for straight pushes. And most importantly, the proposed system's functionality was not evaluated for real world cars, yet. Together with simulated collisions and more complex chains of collision events, these could be subjects of future research.

%%%%%%%%%%%%%%%%%%%%%%%%%%%%%%%%%%%%%

\bibliographystyle{IEEEtran}
\bibliography{refs}



%\end{thebibliography}

\end{document}